\documentclass[journal]{IEEEtran} 
\usepackage{times}
\usepackage{epsfig}
\usepackage{graphics}
\usepackage{amsmath}
\usepackage{float}
\usepackage{amssymb}
\usepackage{subfigure}
\usepackage{footmisc}
\usepackage{indentfirst}
\usepackage{fancyhdr}
\usepackage{multicol}
\usepackage{color}
\usepackage{fullpage,authblk}

\begin{document}
\title{\textbf{Multiple VLAD encoding of CNNs for image classification}}
\author{Qing~Li, Qiang~Peng , Chuan~Yan
\thanks{Qing Li was in the School of Information Science \& Technology, Southwest Jiaotong University,
Chengdu, Sichuan, 610031, P.R. China email: (liqing1988@my.swjtu.edu.cn).}
\thanks{Qiang Peng was in the School of Information Science \& Technology, Southwest Jiaotong University,
Chengdu, Sichuan, 610031, P.R. China email: (pqiang@nec.swjtu.edu.cn).}
\thanks{Chuan Yan was in the School of Information Science \& Technology, Southwest Jiaotong University,
Chengdu, Sichuan, 610031, P.R. China email: (kirin@my.swjtu.edu).}}

\date{}
\maketitle

\centerline{\textbf{Abstract}}
Despite the effectiveness of convolutional neural networks (CNNs) especially in image classification tasks, the effect of convolution features on learned representations is still limited. It mostly focuses on the salient object of the images, but ignores the variation information on clutter and local. In this paper, we propose a special framework, which is the multiple VLAD encoding method with the CNNs features for image classification. Furthermore, in order to improve the performance of the VLAD coding method, we explore the multiplicity of VLAD encoding with the extension of three kinds of encoding algorithms, which are the VLAD-SA method, the VLAD-LSA and the VLAD-LLC method. Finally, we equip the spatial pyramid patch (SPM) on VLAD encoding to add the spatial information of CNNs feature. In particular, the power of SPM leads our framework to yield better performance compared to the existing method.

\begin{IEEEkeywords}
CNNs,VLAD, SPM, Image Classification.
\end{IEEEkeywords}

\section{Introduction}
Deep convolutional neural networks (CNNs), as the powerful image representations for various category-level recognition tasks, have gained a significant amount of applications recently in image classification \cite{krizhevsky2012imagenet}\cite{simonyan2014very} , scene recognition\cite{zhou2014learning} or object detection\cite{girshick2015fast}. On the task of Image classification, how to get the effective representation of the image is very important. CNNs are one of the most notable deep learning approaches and can discover multiple levels of deep representation with the hope that higher-level features can learn from the data of abstractions. Since then, CNNs have consistently been employed in image classification tasks.

The CNNs can learn rich feature representations from images. Meanwhile, they can be considered as a plausible method of remedying the limitations of hand-crafted features. Although a lot of state-of-the art performances have been obtained, CNNs features are still somewhat limited in dealing with the large variation of images. Because CNNs mostly focus on the main objects in the image, they frequently ignore the variation information on clutter and local objects. Therefore, if we directly adopt the CNNs features for image classification task, it should do not produce satisfactory results.

\begin{figure}\centering
  \includegraphics[height=1.5in,width=3.2in]{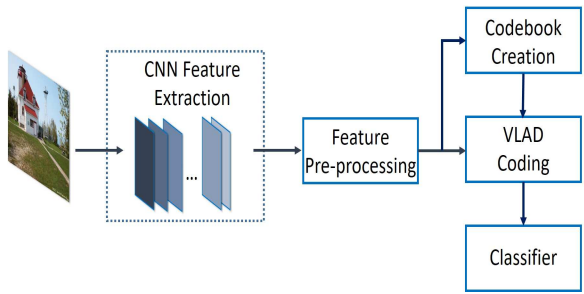}\\
  \caption{The framework of the multiple VLAD encoding method with the CNNs features for image classification. }\label{fig:figure1}
\end{figure}

In this paper, we propose a special framework to solve the problem of CNNs features, which is the multiple VLAD encoding method with the CNNs features for image classification, as shown in Figure \ref{fig:figure1}. The framework contains five steps: 1) CNNs feature extraction, 2) feature pre- processing, 3) codebook creation, 4) VLAD encoding, 5) classification. From this workflow, we can use the VLAD encoding method to capture the local information based on the CNNs features. It not only can keep the global CNNs information of the original image, but also can generate more locally robust representation. Before the CNNs architecture is proposed, the low-level features are represented to the standard hand-crafted are designed, such as SIFT , HOG or local binary patterns (LBP). These feature representations are popularly used on the Bag of Visual Words (BoVW) model with great success for computer vision and image classification tasks \cite{lazebnik2006beyond}\cite{chen2015csift}. However, now the CNNs models are considered to be the primary candidate in feature extraction. The feature pre-processing is important to make the feature representation more stable. For codebook creation, a dictionary is provided and applied to describe the local feature space. The locally aggregated descriptors (VLAD) representation is a kind of efficient super vector encoding method. The VLAD encoding method is used to transform the local features into fixed-size vector representations. Classification is used to evaluate the final classification results. Following this framework, we can generate more locally robust representation, and furthermore, more accurate classification performance.

VLAD encoding is a key pipeline of the framework and can be regarded as a problem of feature mapping. The main issue of VLAD encoding is how to involve the assignment of the local feature descriptors to one or several nearest or a small group of elements on the dictionary. In order to boost the VLAD encoding performance, this paper studies the multiplicity of VLAD and analyze various existing feature coding algorithms. It discovered the underlying relations between these coding algorithms and VLAD. By researching the relation techniques, there developed three kinds of coding methods, VLAD-SA, VLAD-LSA and VLAD-LLC, which embed different kinds of feature encoding approaches into VLAD method.

Moreover, the spatial pyramid matching (SPM), as a traditional model for the BoVW, has been successfully fed into the deep conventional networks. Motivated by the SSP net \cite{he2015spatial} and Fast R-CNN \cite{girshick2015fast}, the spatial information of the local CNNs feature is very important. Therefore, we propose a SPM layer before the VLAD encoding layer in our framework, which called the multiple VLAD encoding method equipped the SPM with CNNs features for image classification. Following this new framework, it can capture the more accurate and robust local CNNs features for the best classification performance.

In summary, the primary contributions of this paper are as followed:

1 we introduce a special framework, which is the multiple VLAD encoding method with the CNNs features for image classification.

2 We explore the multiplicity of VLAD encoding with the extension of several kinds of encoding algorithms, we develop three kinds of coding method, VLAD-SA, VLAD-LSA and VLAD-LLC.

3 We empirically illustrate boosting the performance of classification with VLAD-SA, VLAD-LSA or VLAD-LLC.

4 We equip the SPM on VLAD encoding to add the spatial information of CNNs feature, which can lead a good performance.

\section{Related work}
Reviewed a vast literature on image classification, there has a very challenging problem and gained much attention for many years. One milestone was established by using the low-level features in the BoVW model, such as SIFT (Scale-Invariant Feature Transform), which is a very robust local invariant feature descriptors with respect of geometrical changes \cite{chen2015csift}.

Bag of Visual words (BoVW), as one very classical model of the computer vision society, has been witnessed the popularity and success in image classification \cite{lazebnik2006beyond}\cite{chen2015csift}. BoVW, originated from bag of words model in natural language processing, represents an image as a collection of local features, has been widely used in instance retrieval, scene recognition, and action recognition. Traditionally, vector quantization (hard voting), the most representative encoding method, is one key step in constructing the BoVW model. Over the past several years, a large variety of different feature coding methods have been highly active research areas. For example, in order to solve the L1-norm optimization problem, Wang et al. develop locality-constrained linear coding (LLC) \cite{wang2010locality}. For more large-scale image categorization, super vector encoding methods have obtained the state of-the-art performance in several tasks, especially for the typical methods: Vector of Locally Aggregated Descriptors (VLAD) \cite{jegou2012aggregating}, and Fisher Vector (FV) \cite{simonyan2013deep}. Since super vector encoding methods was shown to achieve the powerful performance on computer vision task \cite{gong2014multi}, we extend to exploring the VLAD encoding methods as an idea to use in our framework.

Recently, the state-of-the-art technique of image classification is the deep convolutional neural networks (CNNs), which is increasingly used in diverse computer vision applications. Generally, CNNs architecture consists of three layers, which is convolutional layers, pooling layers, and fully connected layers. There are many researchers interest in these layers, and enhance the architecture of CNNs followed by changing the specific components in different layers. For example, Gong et al. \cite{gong2014multi} presented a multi-scale orderless pooling scheme (MOP-CNN), which extracts CNN activations for local patches at multiple scale levels, and performs orderless VLAD pooling of these activations at each level separately. Zhun Sun et al. explored the relationship between shape of kernels which define receptive fields (RFs) in CNNs for learning of feature representations and image classification in \cite{sun2016design}.

Because the deep CNNs can be trained in layer-by-layer manner, CNNs are extracted to improve the robustness of learning feature and obtain more high-level image information. Therefore, CNNs as feature extractors are investigated by several authors on all kinds of research areas. Ruobing et al. \cite{wu2015harvesting} presented a novel pipeline built upon deep CNN features for harvesting discriminative visual objects and parts for scene classification. In \cite{laptev2016ti}, Dmitry et al. proposed a deep neural network topology that incorporates a simple to implement transformation-invariant pooling operator (TI-POOLING). Unfortunately, CNNs feature mostly focus on the salient object of the images, but ignores the variation information on clutter and local. To bring in the encoding method can increase the local information based on the CNNs features, especially for using the VLAD encoding methods. Such as the NetVLAD, as a new generalized VLAD layer, was developed by Arandjelovi¡ä c et al. \cite{arandjelovic2016netvlad}. The layer is readily pluggable into any CNN architecture, and learn parameters of the architecture in an end-to-end manner. There are also some works about CNN-based features, which was investigated using the VLAD for image retrieval \cite{yue2015exploiting}, and image captioning task\cite{shin2016dense}. Moreover, the spatial information is essential for improving the classification performance. Because the spatial pyramid matching (SPM) is the popular and computationally efficient extension of the BoVW, it has been successfully fed into the deep conventional networks \cite{he2015spatial}\cite{girshick2015fast}(SSP net and Fast R-CNN).  Therefore, it is very important to build the spatial information for local CNNs feature in this paper.
\section{System Architecture}
\begin{figure}\centering
  \includegraphics[height=1.5in,width=3.2in]{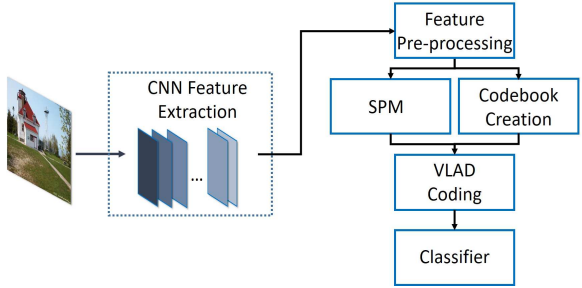}\\
  \caption{The framework of the multiple VLAD encoding method equipped the SPM with the CNNs features for image classification. }\label{fig:figure2}
\end{figure}

In this section, it present the main proposed system architecture, which is multiple VLAD encoding method equipped the SPM with the CNNs features for image classification, seen in Figure \ref{fig:figure2}. And this framework is very similar with the first classification architecture (the multiple VLAD encoding method with the CNNs features for image classification), which is main building the SPM layer before VLAD encoding method. Therefore, we will describe the main pipeline of this framework, which contains: 1) CNNs feature extraction and pre-processing, 2) SPM layer, 3) VLAD encoding.

\subsection{CNNs feature extraction and pre-processing}
The deep convolutional neural networks (CNNs) has been introduced in Section 2, which trained in a robust manner with multiple neural layers. In order to design an end to end CNN architecture, it is more important to consider the CNN model trained on large datasets like ImageNet \cite{krizhevsky2012imagenet}. In this paper, we chose VGG16 \cite{simonyan2014very} and VGG-M \cite{chatfield2014return} as our pre-trained models, which can build the deep CNNs features as well as accelerate the learning process.

Generally, CNNs features mainly focus on the salient object of the images, but ignores the variation information on clutter and local. If we bring in the VLAD encoding method to increase the local information based on the CNNs features, the feature pre-processing is essential to make the features more stable. Because CNNs features are too high-dimensional to be encoded, we use the simple Principle Component Analysis (PCA) method to process these features. The PCA followed by whitening and L2-normalizaition, can be utilized in our experiment and can be furthered enhance the representation of VLAD.

\subsection{SPM layer}
The multiple VLAD encoding method can perform well on the CNNs features for image classification, but this framework still ignores the important spatial information. In order to solve this problem, we build the spatial pyramid matching (SPM) layer before VLAD encoding method. The SPM is a traditional spatial model for the BoVW, and even successfully fed into the deep conventional networks, such as the SSP net and Fast R-CNN.

\begin{figure}\centering
  \includegraphics[height=1in,width=1in]{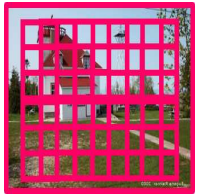}\\
  \caption{The dense grid to capture the patches. }\label{fig:figure3a}
\end{figure}

\begin{figure}\centering
  \includegraphics[height=1.5in,width=3in]{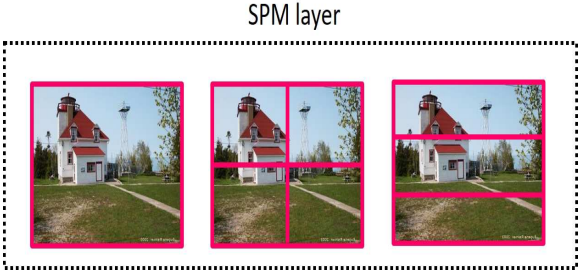}\\
  \caption{A three levels of spatial pyramid matching. }\label{fig:figure3b}
\end{figure}

Inspired by these success algorithms, the image adopts the dense grid to obtain the patches for the SPM layer, as shown in Figure \ref{fig:figure3a}. The size of grid is determined by the numbers of CNNs features. It densely extracted the patches and then pooled on a three levels SPM ($1 \times 1$, $2 \times 2$ and $3 \times 1$), as shown in Figure \ref{fig:figure3b}.

\subsection{VLAD encoding}

In this paper, the key pipeline of the framework is the VLAD encoding, which aggregates the set of local feature descriptors into a fixed-size vector. The VLAD is named the Vector of Locally Aggregated Descriptors and is proposed by J'egou et al. in \cite{jegou2012aggregating}. The same as the BOVW, a dictionary is the indispensable part in VLAD encoding. The idea of the VLAD coding is how to map the local feature descriptors to nearest dictionary, and the step of codebook creation is to generate the dictionary by using the K-means. In the framework of VLAD encoding can be seen in Figure \ref{fig:figure4}.

\begin{figure}\centering
  \includegraphics[height=1in,width=3in]{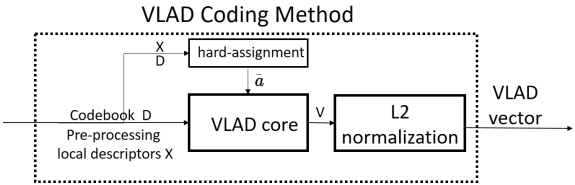}\\
  \caption{The generalized VLAD encoding layer. }\label{fig:figure5}
\end{figure}

Let the $X = [\textbf{x}_{1}, \textbf{x}_{2},\ldots,\textbf{x}_{N}]\in R^{D\times N}$ denote a set of CNNs feature descriptors extracted from an image. The dictionary $D = [\textbf{d}_{1}, \textbf{d}_{2},\ldots,\textbf{d}_{M}]\in R^{D\times M}$ is learned of \emph{m} visual words with the K-means. In the following, the VLAD vector is computed by accumulating the residuals (the vector differences between the assigned descriptor and the nearest the visual words), and it can be represented by $V = [\textbf{v}_{1,1},\ldots,\textbf{v}_{j,m},\ldots,\textbf{v}_{D,M}]$. The element of \emph{V} can be written as follows:

\begin{equation}\label{eq:vladmodel1}
\begin{split}
 & v_{j,m}  ={\sum_{i=1}^{N}\hat{a}_{m}(\textbf{x}_{i})(x_{i}(j),d_{m}(j))}, \\
    \\
 &\hat{a}_{m}(\textbf{x}_{i}) = \left\{\begin{array}{cl}
            1, & \textrm{if $i = \arg\min_{m}|| \textbf{x}-\textbf{d}_{m}||_{2}$},\\
            0, & \textrm{$otherwise$}.
           \end{array}
    \right.
\end{split}
\end{equation}
where $(x_{i}(j)$ and $d_{m}(j))$ respectively denote the $j$-th dimensions of the $i$-th components of descriptor $\textbf{x}$ and the cluster center $\textbf{d}_{m}$. $\hat{a}_{m}(\textbf{x}_{i})$ is the membership of the descriptor $\textbf{x}_{i}$  to the $m$-th visual word, i.e. $\hat{a}_{m}(\textbf{x}_{i})$ is 1 if cluster $\textbf{d}_{m}$ is the closest cluster to descriptor $\textbf{x}_{i}$ and 0 otherwise.The final VLAD vector $V$ is $L2$-normalized for similarity measurement, and expresses as follow:

\begin{equation}\label{eq:vladl2}
V
= \begin{bmatrix}
 \frac{\textbf{v}_{1,1}}{||\textbf{v}_{1,1}||_{2}};&\dots;&
 \frac{\textbf{v}_{j,m}}{||\textbf{v}_{j,m}||_{2}};&\dots;&
 \frac{\textbf{v}_{D,M}}{||\textbf{v}_{D,M}||_{2}}\\
\end{bmatrix}.
\end{equation}

The binary assignment weight indicating $\hat{a}_{m}(\textbf{x}_{i})$ is obtained by the hard assignment in the original VLAD encoding. In this paper, the main issue is how to boost the VLAD encoding performance. Motived by the hard assignment method, it sniffs out the underlying relations between the coding algorithms and VLAD encoding. It means to research the assignment of the local feature descriptors to one visual word or several nearest visual words or a small group of elements on the dictionary. Therefore, we explore several practical encoding techniques, and design three kinds of feature coding approaches to embed into the VLAD encoding, seen in Figure \ref{fig:figure5}.

\begin{figure}\centering
  \includegraphics[height=1.5in,width=3in]{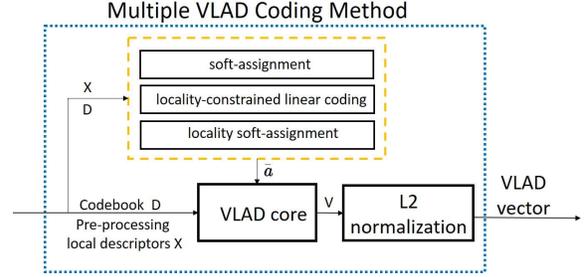}\\
  \caption{The expanding VLAD encoding layer embed with three kinds of feature coding approaches. }\label{fig:figure6}
\end{figure}

The first is the VLAD-SA method, which is used the soft assignment to replace the hard assignment. The soft assignment uses the kernel function of distance as the coding representation. Then, the binary assignment weight indicating $\hat{a}_{m}(\textbf{x}_{i})$ can be written as

\begin{equation}\label{eq:SCmodel1}
    \hat{a}_{m}(\textbf{x}_{i}) =\frac{exp(-\beta || \textbf{x}-\textbf{d}_{m}||^{2}_{2}}{\sum_{n=1}^{M}exp(-\beta ||\textbf{x}-\textbf{d}_{n}||^{2}_{2}}
 \end{equation}
where $\beta$ is a smoothing factor controlling the softness of the assignment. Note that multiple visual words $\textbf{d}_{k}$ are employ for coding with the descriptor $\textbf{x}$.

The second is the VLAD-LSA method, which is used the localized soft assignment coding. Expanding the manifold structure into the descriptor space, it only considered the $k$ nearest visual words are used to assign with the descriptor:

\begin{equation}\label{eq:SCmodel1}
\begin{split}
   & \hat{a}_{m}(\textbf{x}_{i})  =\frac{\hat{D}(\textbf{x},\textbf{d}_{m})exp(-\beta || \textbf{x}-\textbf{d}_{m}||^{2}_{2}}{\sum_{n=1}^{M}\hat{D}(\textbf{x},\textbf{d}_{m})exp(-\beta || \textbf{x}-\textbf{d}_{n}||^{2}_{2}}, \\
    \\
    & \hat{D}(\textbf{x},\textbf{d}_{m})= \left\{\begin{array}{cl}
            1 & \textrm{if $d_{m} \in NN_{K}(\textbf{x})$}\\
            0 & \textrm{$otherwise$}
           \end{array}
    \right.
\end{split}
\end{equation}
where $\hat{D}(\textbf{x},\textbf{d}_{m})||\textbf{x}-\textbf{d}_{m}||^{2}_{2}$ is the localized version of the original distance $|| \textbf{x}-\textbf{d}_{m}||^{2}_{2}$, and $\hat{D}(\textbf{x},\textbf{d}_{k})$ is the indicator function to denote the $K$-nearest neighborhood of $\textbf{x}_{i}$.

The last one is the VLAD-LLC method, which is used the Locality-constrained Linear Coding (LLC) to enforce locality instead of sparse constraint. It leads to smaller coefficient for the basis vectors farther away from the local feature $\textbf{x}_{i}$. The LLC coding coefficient is obtained by the criteria:

\begin{equation}\label{eq:LLCmodel1}
\begin{split}
    & \hat{a}_{m}(\textbf{x}_{i}) =  \arg\min_{\hat{a}\in R^{M}}\|\textbf{x}_{i}-D\hat{a}\|^{2}+\lambda\|\textbf{s}_{i}\odot \hat{a}\|^{2} \\
    \\
    & s.t. 1^{T}\hat{a}_{m}(\textbf{x}_{i})=1,\forall i
\end{split}
\end{equation}
where $\odot $ denotes the element-wise multiplication, and $s_{i}$ is the locality adaptor that ensures weight for each basis vector are the proportional with its similarity to the input descriptor $\textbf{x}_{i}$:
\begin{equation}\label{eq:LLCmodel2}
    s_{i}=\exp[\frac{dist(\textbf{x}_{i},D)}{\sigma}]
\end{equation}
where $dist(\textbf{x}_{i},D)=[dist(\textbf{x}_{i},\textbf{d}_{1}),dist(\textbf{x}_{i},\textbf{d}_{2}),\ldots,$ $dist(\textbf{x}_{i},\textbf{d}_{M})]$, and $dist(x_{i},d_{m})$ is the Euclidean distance between $x_{i}$ and $d_{m}$.$\sigma$ is the parameter adjusting the weight decay speed for the locality adaptor $s_{i}$.To further improve the computational, an approximated LLC can be used in practice. It simply uses the $K$ nearest basis vectors of $\textbf{x}$ to minimize the first term and ignore the second term in Eq. (\ref{eq:LLCmodel1}) by solving a much smaller linear system. In this condition, the code coefficients is the selected $k$ basis vectors and others are simply set to zero.

\section{Experiments}

To illustrate the performance of the multiple VLAD encoding methods, an empirical research is presented in this section. First, we introduce databases and settings. Then we verify the effectiveness of three VLAD encoding methods with CNN features, followed with the results analysis. Finally, we explore VLAD encoding jointed with the different spatial blocks of spatial pyramid patch, and evaluate the performance of this framework.

\subsection{Settings}
\emph{\textbf{Datasets}}
We evaluate the performance of our framework on two datasets: Caltech-101 and Caltech-256. The Caltech-101 dataset contains 9144 images with a variety of object classes and a background class. All of object classes includes animals, flowers, cars, etc. In this experiment, it sets 30 images per class for training in the whole dataset and the rest images for testing.
 The Caltech-256 dataset contains 30607 images of 256 classes. Compared with Caltech-101 dataset, it represents a more formidable variability in location, background, image size and lighting conditions. Moreover, the minimum number of images in any category are increased from 31 to 80. In this experiment, it sets 30 and 60 images per class for training in the whole dataset and the rest images for testing.

\emph{\textbf{Deep learning model}}
 The VGG-M and VGG16 are choosen as our pre-trained models, which can build the deep CNNs feature maps based representation. VGG-M is characterized by the decreased stride and smaller receptive field of the first convolutional layer \cite{chatfield2014return}. The CNN-M contains 5 convolutional layers, and three fully-connected layers. In VGG-M, especially for the conv2 uses lager stride to keep the computation time reasonable. The VGG-M is a simple and fast model for the evaluation of CNN-based method for image classification. The VGG 16 is designed to increase depth of the network by using an architecture with very small coevolution filters in all layers \cite{simonyan2014very}. It contains 16 weight layers including 13 convolutional layers with filtersize of $3 \times 3$, and 3 fully-connected layers. In the VGG 16, all convolutional layers are divided into 5 groups, each of which has pooling and downsampling layers.

\subsection{Implementation details}
The CNNs feature descriptors are extracted on the VGG-M and VGG16 models. In these models, they use the conv5 layer of CNNs as the feature extractor, and the pixels of conv5 feature maps as the local features descriptors to encode with the VLAD method. After the extracted the CNNs feature, all feature descriptors need PCA processing and whitening. Then, dictionary is learned from a subset of CNNs feature descriptors. Dimensionality of the dictionary is fixed to 64. The effectiveness of three VLAD encoding methods are verified on datasets and compared with the benchmark of CNNs model on classification. In order to increase the spatial information, the SPM is added in this framework. It densely extract the local patches with the corresponding conv5 feature mapped for the spatial pyramid matching. As for SPM, it is divided in $1 \times 1$, $2 \times 2$ and $3 \times 1$ grids, $1 \times 1$, $2 \times 2$ and $1 \times 3$ grids and $1 \times 1$, $2 \times 2$ and $4 \times 4$ grids. At last, we examine the VLAD encoding jointed with the different spatial blocks of spatial pyramid patches, and contrast the performance of the VLAD encoding with and without the SPM.

\subsection{Evaluation of CNN features with VLAD encoding}
Here we use the cov5 and softmax layer of the VGG-M model, and the $cov5_4$ and softmax layer of the VGG16 model. After the cov5 or $cov5_4$ layer, the VLAD encoding is utilized to generate a compact and efficient representation, and explored to be the multiplicity of VLAD encoding with the extension of several kinds of encoding algorithms. Here it presents three kinds of coding methods, which is the VLAD-SA, VLAD-LSA and VLAD-LLC coding methods. The classical full-connected CNNs features mainly pool by max pooling, they provide a comparison of different kinds of coding schemes as well as result for the final framework.
Seen from Table 1 and Table 2, they present our different kinds of VLAD encoding method has been outperform the CNNs features based with single model. The results in Table 1 and Table 2 are reported our proposed encoding methods can improve the performance of classification. Table 1 shows the best performance is $83.72\%$ on VGG-M model and $89.23\%$ on VGG16 model by using the VLAD-LLC coding method on Caltech101 datasets. Table 2 shows the best performance is $64.85\%$ on VGG-M model and $74.85\%$ on VGG16 model by using the VLAD-SA coding method on Caltech256 datasets. In our experiment, if we set 30 images per class for training on Caltech256 datasets, we will find out that the best one is VLAD-LLC coding method, the second is the VLAD-LSA coding method. However, when the training is employed 60 images per class, the VLAD-SA coding method encode more feature information and obtain the slight improvement of the result.

\begin{table*}
\centering
\caption{The comparison of different kinds of VLAD encoding method on Caltech-101(Training images 30)}
\label{tab:table1}
\begin{tabular}{|c|c|c|c|c|c|c|c|}
\hline \scriptsize Model & \scriptsize CNN & \scriptsize Ours (CNN+VLAD) &\scriptsize Ours(CNN+VLAD-SC) &\scriptsize Ours(CNN+VLAD-LSA) &\scriptsize Ours(CNN+VLAD-LLC)&\scriptsize CSIFT \cite{chen2015csift} &\scriptsize LLC \cite{wang2010locality} \\
\hline \scriptsize VGG-M &\scriptsize 66.05 &\scriptsize 78.41 &\scriptsize 79.53 &\scriptsize 83.43 &\scriptsize \textbf{83.72} &\scriptsize -- &\scriptsize--\\
 \hline \scriptsize VGG16 &\scriptsize 71.77 &\scriptsize 84.19 &\scriptsize 87.57 &\scriptsize 88.85 &\scriptsize \textbf{89.23} &\scriptsize -- &\scriptsize --\\
  \hline \scriptsize No Model &\scriptsize -- &\scriptsize -- &\scriptsize -- &\scriptsize -- &\scriptsize -- &\scriptsize  72.39 &\scriptsize 73.44 \\
\hline
\end{tabular}
\end{table*}


\begin{table*}
\centering
\caption{The comparison of different kinds of VLAD encoding method on Caltech-256(Training images 60)}
\label{tab:table2}
\begin{tabular}{|c|c|c|c|c|c|c|c|}
\hline \scriptsize Model & \scriptsize CNN & \scriptsize Ours (CNN+VLAD) &\scriptsize Ours(CNN+VLAD-SC) &\scriptsize Ours(CNN+VLAD-LSA) &\scriptsize Ours(CNN+VLAD-LLC)&\scriptsize CSIFT \cite{chen2015csift} &\scriptsize LLC \cite{wang2010locality} \\
\hline \scriptsize VGG-M &\scriptsize 53.18 &\scriptsize 55.36 &\scriptsize \textbf{64.85} &\scriptsize 63.0 &\scriptsize 63.17 &\scriptsize -- &\scriptsize -- \\
 \hline \scriptsize VGG16 &\scriptsize 55.08 &\scriptsize 67.08 &\scriptsize \textbf{74.85} &\scriptsize 73.90 &\scriptsize 74.25  &\scriptsize -- &\scriptsize -- \\
 \hline \scriptsize No Model &\scriptsize -- &\scriptsize -- &\scriptsize -- &\scriptsize -- &\scriptsize -- &\scriptsize  41.31 &\scriptsize 47.68 \\
\hline
\end{tabular}
\end{table*}

\subsection{Evaluation of CNN features based on VLAD encoding with SPM}

In this section, it equips the spatial pyramid patch (SPM) combined with different kinds of VLAD encoding method. The method of CNN is used the stand CNN framework from the model on \cite{chatfield2014return}\cite{simonyan2014very}, and obtained the results without finetuning. The proposed methods are followed these benchmarks to set the parameters of the experiment. Table 3 summarizes the results from our framework with SPM. It present the best performance is $87.49\%$ on VGG-M model and $92.54\%$ on VGG16 model by using the VLAD-LLC coding method on Caltech101 datasets. However, it chooses the 60 images per class on Caltech256 datasets, the system cannot get enough memory to training the classifier on the VLAD-SA encoding method. This is because of the VLAD-SA coding method can reduce the information loss during encoding, the final vector need more memory space to save. Therefore, it shows the best performance is $68.02\%$ on VGG-M model and $76.46\%$ on VGG16 model by using the VLAD-LLC coding method.

\begin{table*}
\centering
\caption{The comparison of different kinds of VLAD encoding method with SPM}
\label{tab:table3}
\begin{tabular}{|c|c|c|c|c|c|}
\hline \multicolumn{6}{|c|}{Comparison On Caltech-101 (Training images 30)} \\
\hline \scriptsize Model & \scriptsize CNN  & \scriptsize Ours (CNN+VLAD) &\scriptsize Ours(CNN+VLAD-SC) &\scriptsize Ours(CNN+VLAD-LSA) &\scriptsize Ours(CNN+VLAD-LLC) \\
\hline \scriptsize VGG-M &\scriptsize 66.05 \cite{chatfield2014return} &\scriptsize 84.17 &\scriptsize 84.20 &\scriptsize 87.15 &\scriptsize \textbf{87.49} \\
 \hline\scriptsize VGG16 &\scriptsize 71.77 \cite{simonyan2014very} &\scriptsize 89.31 &\scriptsize 90.51 &\scriptsize 92.54 &\scriptsize \textbf{92.54}  \\
 \hline \multicolumn{6}{|c|}{Comparison On Caltech-256 (Training images 60)} \\[3pt]
\hline \scriptsize Model & \scriptsize CNN & \scriptsize Ours (CNN+VLAD) &\scriptsize Ours(CNN+VLAD-SC) &\scriptsize Ours(CNN+VLAD-LSA) &\scriptsize Ours(CNN+VLAD-LLC)\\
\hline \scriptsize VGG-M &\scriptsize 53.18 \cite{chatfield2014return} &\scriptsize 62.79 &\scriptsize -- &\scriptsize 67.71 &\scriptsize \textbf{68.02} \\
 \hline \scriptsize VGG16 &\scriptsize 55.08 \cite{simonyan2014very} &\scriptsize 72.53 &\scriptsize -- &\scriptsize 76.41 &\scriptsize \textbf{76.46}  \\
  \hline
\end{tabular}
\end{table*}

Compared the VLAD encoding with and without SPM layer, it obviously illustrates that adding the spatial pyramid can improve the overall performance of classification. In order to research the spatial pyramid, this paper presents three kinds of spatial regions, which are divided in $1 \times 1$, $2 \times 2$ and $3 \times 1$ grids, $1 \times 1$, $2 \times 2$ and $1 \times 3$ grids and $1 \times 1$, $2 \times 2$ and $4 \times 4$ grids. The results of this experiments over Caltech 256 datasets are shown in Table 4, and 30 images per class are set for training. From these results, our framework is not sensitive to the spatial regions. Nonetheless, we still evaluate these approaches which are the best choice. In these cases, the best performance and the pooling time are considered to be the evaluation criteria. Seen from the Table 4, it present the shows the best performance is $63.0\%$ on VGG-M model and $72.50\%$ on VGG16 model by using the VLAD-LLC coding method with the first SPM division. Especially for the third one, it has 21 pathes and need more time to pool the features from arbitrary windows on feature maps. Therefore, the first one is the best division which maintains the high performance when using the lower pooling time.

\begin{table*}
\centering
\caption{The comparison of three kinds of spatial pyramid regions}
\label{tab:table4}
\tabcolsep0.25in
\renewcommand{\arraystretch}{1.2}
\begin{tabular}{|c|c|c|c|c|}
\hline \multicolumn{5}{|c|}{Comparison On Caltech-256 (Training images 30)} \\[3pt]
\hline \multicolumn{5}{|c|}{SPM divided in $1 \times 1$, $2 \times 2$ and $3 \times 1$ grids} \\[3pt]
\hline \scriptsize Model & \scriptsize CNN+VLAD &\scriptsize CNN+VLAD-SC &\scriptsize CNN+VLAD-LSA &\scriptsize CNN+VLAD-LLC\\
\hline \scriptsize VGG-M &\scriptsize 57.34 &\scriptsize 58.31 &\scriptsize 62.97 &\scriptsize \textbf{63.0} \\
 \hline \scriptsize VGG16 &\scriptsize 67.03 &\scriptsize 69.39 &\scriptsize 72.36 &\scriptsize \textbf{72.50}  \\
\hline \multicolumn{5}{|c|}{SPM divided in $1 \times 1$, $2 \times 2$ and $1 \times 3$ grids} \\[3pt]
\hline \scriptsize VGG-M &\scriptsize 56.72 &\scriptsize 58.62 &\scriptsize 62.54 &\scriptsize 62.54 \\
 \hline \scriptsize VGG16 &\scriptsize 66.55 &\scriptsize 69.37 &\scriptsize 71.40 &\scriptsize 71.41  \\
 \hline \multicolumn{5}{|c|}{SPM divided in $1 \times 1$, $2 \times 2$ and $4 \times 4$ grids} \\[3pt]
 \hline \scriptsize VGG-M &\scriptsize 56.25 &\scriptsize -- &\scriptsize 61.50 &\scriptsize 61.43 \\
 \hline \scriptsize VGG16 &\scriptsize 66.80 &\scriptsize -- &\scriptsize 71.48 &\scriptsize 71.46  \\
\hline
\end{tabular}
\end{table*}
\section{Conclusion}
In this paper, we first designed a special image classification framework that is the multiple VLAD encoding method with the CNNs features. Our framework significantly improves over the traditional CNNs model on image classification. We then explored the multiplicity of VLAD encoding with the extension of several kinds of encoding algorithms, so that these can boost the performance of VLAD coding method. We offered three kinds of coding methods embed into VLAD encoding, they are called VLAD-SA, VLAD-LSA and VLAD-LLC. We verified the effectiveness of those methods for the task of image classification, the VLAD-LLC coding method is the best one on Caltech101 datasets and the VLAD-SA coding method is the best on Caltech256 datasets. This is because of the VLAD-SA coding method can reduce the information loss during encoding, when the final vectors become larger and need more memory space to save. Finally, we combined the spatial pyramid patch (SPM) with VLAD encoding to add the spatial information of CNNs feature. In our experiments show that the proposed framework with SPM achieve the better classification accuracy over the traditional CNNs model. Following those research on the multiple VLAD encoding method with the CNNs features, the CNNs feature based the feature encoding representation methods can lead more satisfactory performance than traditional CNNs architecture. In this experiment,we use the dictionary is only set 64 dim. If we improve the size of the dictionary, our method can obtain more better performances. However, the memory space is not enough to compare the size of the dictionary, we will do this research in next research. In the future, we will integrate these feature encoding approaches into the CNNs framework, and explore these new framework to implement in more homogeneous applications.

\section*{Acknowledgments}
This work is supported by the Fundamental Research Funds for Central Universities(No.2062015YXZT11).

\bibliographystyle{IEEEtran}
\bibliography{ref}

\end{document}